\documentclass{article}

\usepackage[preprint]{neurips_2026}

\usepackage[utf8]{inputenc}
\usepackage[T1]{fontenc}
\usepackage{hyperref}
\usepackage{url}
\usepackage{booktabs}
\usepackage{amsfonts}
\usepackage{amsmath}
\usepackage{amssymb}
\usepackage{graphicx}
\usepackage{nicefrac}
\usepackage{microtype}
\usepackage{xcolor}
\usepackage{float}

\title{Learning a Normal World Model for Few-Shot Boundary-Calibrated Abnormality Detection}

\author{
Weizhi Nie, Weichao Liu, Weijie Wang, and Yuting Su\\
Tianjin University
}

\newcommand{\Eall}{E_{\mathrm{all}}}
\newcommand{\Edyn}{E_{\mathrm{dyn}}}
\newcommand{\Ehyp}{E_{\mathrm{hyp}}}
\newcommand{\Eman}{E_{\mathrm{man}}}
\newcommand{\RUL}{\mathrm{RUL}}

\begin{document}

\maketitle

\begin{abstract}
Abnormality detection in complex systems faces two practical barriers: abnormal labels are scarce, and binary labels do not quantify how far an event has departed from normal behavior.
We study a normal-world modeling formulation for this setting.
Instead of learning a large and incomplete space of abnormal classes, the model learns the normal world from abundant normal events and uses a few abnormal examples only to calibrate the boundary of normality.
We instantiate this idea as a Hypergraph Entropic Normal-World Model.
The model represents multivariate sensor windows as context-conditioned hypergraphs, where hyperedges capture high-order relations among groups of variables.
It then defines abnormality by an entropy-aware normal-world energy that combines temporal prediction surprise, hypergraph consistency surprise, and latent normal-manifold departure.
On the NASA C-MAPSS turbofan degradation benchmark, the proposed full energy achieves strong zero-shot and few-shot performance across all four subsets and reaches AUROC 0.9983 on FD004, the most complex setting with multiple operating conditions and fault modes.
Beyond standard detection metrics, we introduce mechanistic validation tests to probe whether the energy encodes normal-world structure rather than a superficial input-output mapping.
The learned energy accepts unseen healthy engines, increases along degradation trajectories, and sharply penalizes context-mismatched cross-variable coupling breaks.
These results suggest that normal-world energy can serve as an anomaly score, a graded risk measure, and a testable representation of normal system behavior under severe abnormal-label scarcity.
\end{abstract}

\begin{figure}[t]
\centering
\IfFileExists{figures/motivation.png}{
\includegraphics[width=\linewidth]{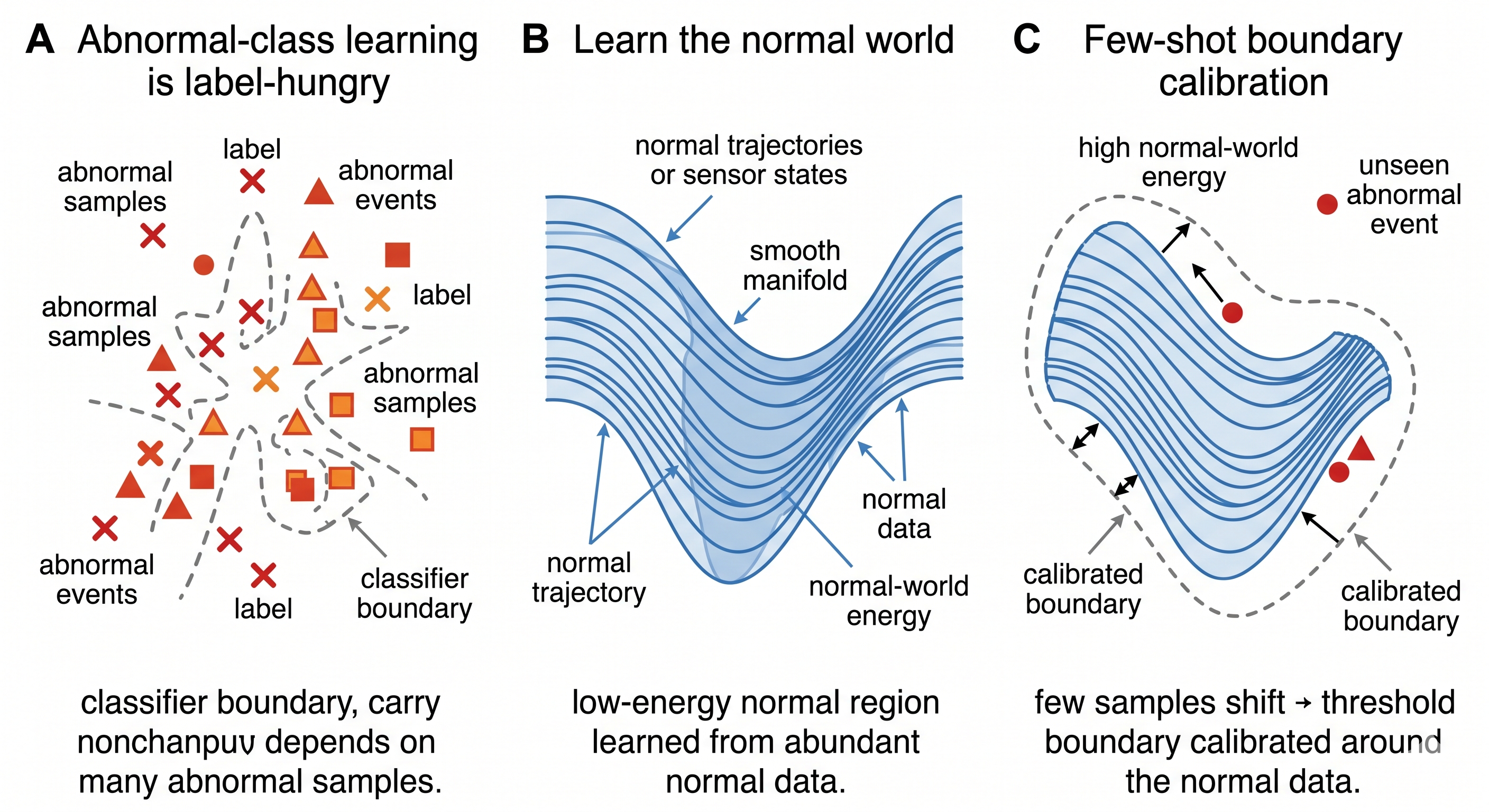}
}{
\fbox{\parbox{0.92\linewidth}{\centering
Place the motivation figure at \texttt{figures/motivation.png}.}}
}
\caption{Motivation. Abnormal-class learning is label hungry, while a normal world model first learns the low-energy normal region from abundant normal events and then uses a few abnormal examples only to calibrate the boundary of normality.}
\label{fig:motivation}
\end{figure}

\section{Introduction}

Detecting abnormal events is central to industrial fault diagnosis, clinical monitoring, and cyber-physical systems.
In these settings, the goal is not only to classify a failure after it appears.
We also want to know whether a current event is still compatible with normal operation.
Two barriers make this difficult.
The first barrier is label scarcity.
Real failures are rare, costly to collect, and diverse.
An engine can fail through many mechanisms, and a patient can deteriorate through many physiological pathways.
For a new fault mode, a practitioner may have only one or two abnormal examples.
This is too little for conventional supervised training.

The second barrier is scoring.
Even when a few abnormal examples exist, a binary label does not say \emph{how abnormal} an event is.
It gives no graded normality score, no risk margin, and no stable boundary across operating contexts.
This problem is acute in multivariate systems.
Abnormality may not appear as a large deviation in one sensor.
In a turbofan engine, temperature, pressure, speed, and flow are coupled by the physical state.
Each sensor can look plausible on its own, while the joint configuration no longer matches any normal regime.
The challenge is therefore twofold: abnormal labels are scarce, and departure from normality needs a principled quantitative score.

These barriers motivate a simple change in viewpoint.
If failures cannot be learned from scarce abnormal examples, we learn normality from abundant normal data.
Normal operation is usually plentiful.
Machines spend most of their lifetime in healthy regimes, and many clinical observations are collected before severe deterioration.
We therefore ask whether a model can learn a \emph{normal world model}: how normal events evolve, how variables co-vary, and how these relations change with context.
After this world is learned, abnormality can be treated as departure from the normal world rather than membership in a predefined fault class.
Figure~\ref{fig:motivation} illustrates this view.

This viewpoint also gives a scoring principle.
Normal and abnormal events should have different information-theoretic surprise under the learned normal world.
Events that follow normal dynamics and high-order relations should have low uncertainty and low energy.
Events that violate those relations should have higher surprise.
We therefore define an entropy-aware normal-world energy.
The energy measures how difficult an event is to explain under the learned normal model.
It can be used as a normality score, a risk margin, and a decision statistic after few-shot boundary calibration.

We instantiate this idea with a Hypergraph Entropic Normal World Model.
Given a multivariate event window, the model treats sensors as nodes and constructs a context-conditioned hypergraph whose hyperedges represent high-order sensor relations.
The model learns three complementary aspects of the normal world from normal windows.
The dynamic branch predicts future sensor states and measures temporal surprise.
The hypergraph consistency branch reconstructs each sensor through cross-variable hypergraph messages and measures violation of high-order relations.
The latent manifold branch measures the distance between the event representation and the normal latent region.
These terms are combined into an entropic normal-world energy.
During few-shot calibration, the model parameters remain fixed; one or two abnormal examples only adjust the decision threshold around the learned normal region.

This paper makes three contributions:
\begin{itemize}
    \item We propose a normal world modeling framework for few-shot abnormality detection. Abundant normal events learn the normal operating world, while scarce abnormal events only calibrate the boundary.
    \item We introduce an entropy-aware normal-world energy for quantitative abnormality evaluation. It combines temporal surprise, hypergraph consistency surprise, and latent manifold departure into a graded normality score.
    \item We evaluate the method on the NASA C-MAPSS turbofan degradation benchmark \citep{saxena2008cmapss}. The results show strong zero-shot and few-shot performance. Mechanistic tests further support that the learned energy captures normal-world structure.
\end{itemize}

\section{Related Work}

\paragraph{Normality modeling and one-class anomaly detection.}
Anomaly detection has a long history in statistics and machine learning \citep{chandola2009survey,pang2021review}.
Classical one-class methods estimate the support of normal data or isolate low-density samples.
Examples include one-class SVM \citep{scholkopf2001support}, support vector data description \citep{tax2004svdd}, and isolation forest \citep{liu2008isolation}.
Deep one-class methods replace hand-crafted features with learned representations.
Representative examples include Deep SVDD \citep{ruff2018deep}, DAGMM \citep{zong2018dagmm}, DROCC \citep{goyal2020drocc}, and FCDD \citep{liznerski2021fcdd}.
These methods are important predecessors because they exploit the asymmetry between abundant normal data and scarce abnormal data.
Most of them learn a compact normal region or a reconstruction/density score.
Our goal is different.
We learn a normal \emph{world model} that combines temporal dynamics, high-order cross-variable consistency, contextual dependence, and boundary calibration.

\paragraph{Deep multivariate time-series anomaly detection.}
Modern multivariate time-series anomaly detection methods often use prediction or reconstruction error as the anomaly score.
Representative recurrent and generative models include LSTM encoder-decoder anomaly detection \citep{malhotra2016lstm}, Telemanom \citep{hundman2018telemanom}, MSCRED \citep{zhang2019mscred}, MAD-GAN \citep{li2019madgan}, TadGAN \citep{geiger2020tadgan}, OmniAnomaly \citep{su2019omni}, and USAD \citep{audibert2020usad}.
More recent methods use attention and Transformer architectures for long-range temporal patterns.
This group includes MTAD-GAT \citep{zhao2020mtadgat}, the Anomaly Transformer \citep{xu2022anomaly}, TranAD \citep{tuli2022tranad}, DTAAD \citep{yu2023dtaad}, and DCdetector \citep{yang2023dcdetector}.
Broader representation models also show the value of strong temporal features.
Examples include TimesNet \citep{wu2023timesnet}, PatchTST \citep{nie2023patchtst}, and MOMENT \citep{goswami2024moment}.
Related temporal event modeling has also been explored in biomedical image sequences, including multi-grained random fields and explicit temporal modeling for mitotic event detection \citep{liu2017multigrained,nie2017mitotictemporal}.
Our method differs from these approaches in two ways.
First, it treats anomaly scoring as normal-world violation rather than only forecast/reconstruction residual.
Second, it is designed for few-shot boundary calibration: the representation and energy are learned from normal data, while scarce abnormal events only adjust the threshold.

\paragraph{Graph and hypergraph modeling for coupled variables.}
Multivariate time series often contain structured variable interactions.
Graph-based forecasting and anomaly detection methods model pairwise sensor relations with spatial-temporal graph networks \citep{wu2019graphwavenet,wu2020mtgnn,cao2020stemgnn}.
For anomaly detection, MTAD-GAT \citep{zhao2020mtadgat}, GDN \citep{deng2021gdn}, and GLUE \citep{ray2021glue} show that relational structure can improve detection of multivariate failures.
Graph-based relation modeling has also been used in domain-specific fault and anomaly settings, including rolling bearing fault diagnosis and ship behavior anomaly detection \citep{yin2023bearingfault,ma2026shipanomaly}.
However, pairwise graphs can be limiting when the normal state depends on high-order relations among groups of sensors.
Hypergraph learning provides a natural representation for group-wise interactions.
Its foundations include hypergraph embedding \citep{zhou2006hypergraphs}.
Neural variants include HGNN \citep{feng2019hgnn}, Hyper-SAGNN \citep{zhang2020hypersagnn}, AllSet \citep{chien2022allset}, and equivariant hypergraph neural networks \citep{kim2022ehnn}.
Structured relation learning has also been studied through hyper-clique graph matching, subgraph learning, and hierarchical graph structure learning \citep{nie2018hyperclique,nie2020subgraph,su2018hierarchicalgraph}.
Our model follows this high-order relational view.
It also makes the hypergraph context-conditioned, so valid sensor relations can change across operating regimes.

\paragraph{Few-shot and weakly supervised abnormality detection.}
Because real abnormal samples are rare, several works study anomaly detection with weak, partial, or scarce abnormal supervision.
Deviation networks use a small number of labeled anomalies to learn a deviation score \citep{pang2019devnet}.
Deep SAD extends deep one-class learning to semi-supervised anomaly detection \citep{ruff2020deepsad}.
Other weakly supervised and deep anomaly detection surveys emphasize the difficulty of abnormal-label coverage and evaluation bias \citep{pang2021review}.
These methods typically use abnormal examples to shape the representation or anomaly scoring function.
In contrast, our setting uses abnormal examples only for boundary calibration.
This separation is important: the normal world model can be trained without assuming that the available abnormal samples span the possible failure modes.

\paragraph{Entropy, energy, and world models.}
Our scoring function is related to information-theoretic and energy-based views of learning.
Shannon entropy provides the classical measure of uncertainty \citep{shannon1948communication}, and energy-based models define compatibility through a scalar energy \citep{lecun2006tutorial}.
Score matching \citep{hyvarinen2005score} and noise-contrastive estimation \citep{gutmann2010nce} provide two ways to learn energy functions.
Joint energy-based models offer another interpretation \citep{grathwohl2020jem}.
In anomaly and out-of-distribution detection, energy scores have also been used to quantify whether a test input is compatible with the training distribution \citep{hendrycks2017baseline,liu2020energyood}.
Our work uses this idea in a structured time-series setting: energy is not only a density-like scalar, but a combination of predictive surprise, hypergraph consistency surprise, and manifold departure.

The term world model is widely used for models that capture the dynamics and latent structure of an environment \citep{ha2018worldmodels,hafner2019dreamer,hafner2023dreamerv3}.
Recent self-supervised predictive architectures make a related point.
I-JEPA and joint-embedding prediction methods learn useful latent regularities without direct labels \citep{assran2023ijepa,bardes2024vjepa}.
Our paper adopts the world-model perspective for abnormality detection rather than control or visual representation learning.
The normal world model is learned from normal system behavior, and abnormality is evaluated as information-theoretic departure from this learned world.

\section{Problem Formulation}

Let \(x_t=[s_{t-W+1},\ldots,s_t]\in \mathbb{R}^{W \times D}\) denote a multivariate event window with length \(W\) and \(D\) sensor variables, where \(s_t\in\mathbb{R}^{D}\) is the sensor state at time \(t\).
Let \(c_t \in \mathbb{R}^{C}\) denote the operating context, such as load, speed setting, environmental condition, or a learned context representation.
We assume that the observed normal events are sampled from an unknown normal-world process \(P_N(x,c)\).
The training set contains many normal windows,
\begin{equation}
\mathcal{D}_N=\{(x_i,c_i)\}_{i=1}^{N},
\end{equation}
while only a few abnormal calibration windows are available,
\begin{equation}
\mathcal{D}_A^K=\{(x_i^-,c_i^-)\}_{i=1}^{K}, \quad K \in \{1,2,4,8\}.
\end{equation}

The task is not to learn a discriminative classifier over all abnormal classes.
Such a classifier would require abnormal samples that cover the space of possible faults, which is unrealistic in the intended few-shot setting.
Instead, we learn a normal world model \(M_\Theta\) from \(\mathcal{D}_N\).
The model induces a scalar energy function
\begin{equation}
E_\Theta(x,c):\mathbb{R}^{W\times D}\times\mathbb{R}^{C}\rightarrow \mathbb{R},
\end{equation}
where lower energy indicates stronger compatibility with the learned normal world.
This energy is used for three related purposes:
\begin{equation}
\mathrm{Normality}(x,c)=\exp[-E_\Theta(x,c)],
\end{equation}
\begin{equation}
\mathrm{RiskMargin}(x,c;\tau)=E_\Theta(x,c)-\tau,
\end{equation}
and
\begin{equation}
\hat{y}=\mathbb{I}[E_\Theta(x,c)>\tau_K].
\end{equation}

The central question is therefore how to construct \(E_\Theta\) so that it measures departure from normal-world structure rather than only marginal distributional distance.
We use an information-theoretic view.
If \(q_\Theta(\cdot|x,c)\) denotes the predictive or consistency distribution learned from normal events, then the information surprise of an observation \(y\) is
\begin{equation}
\mathcal{I}_\Theta(y|x,c)=-\log q_\Theta(y|x,c).
\end{equation}
For normal events, the observation should be easy to encode under the normal model and should have low surprise.
For abnormal events, the model should either make a large prediction error, assign high uncertainty, or fail to explain the observation through normal cross-variable relations.
Thus, our energy is designed as an approximation of normal-world information surprise.

\section{Method}

\begin{figure}[t]
\centering
\IfFileExists{figures/workflow.png}{
\includegraphics[width=\linewidth]{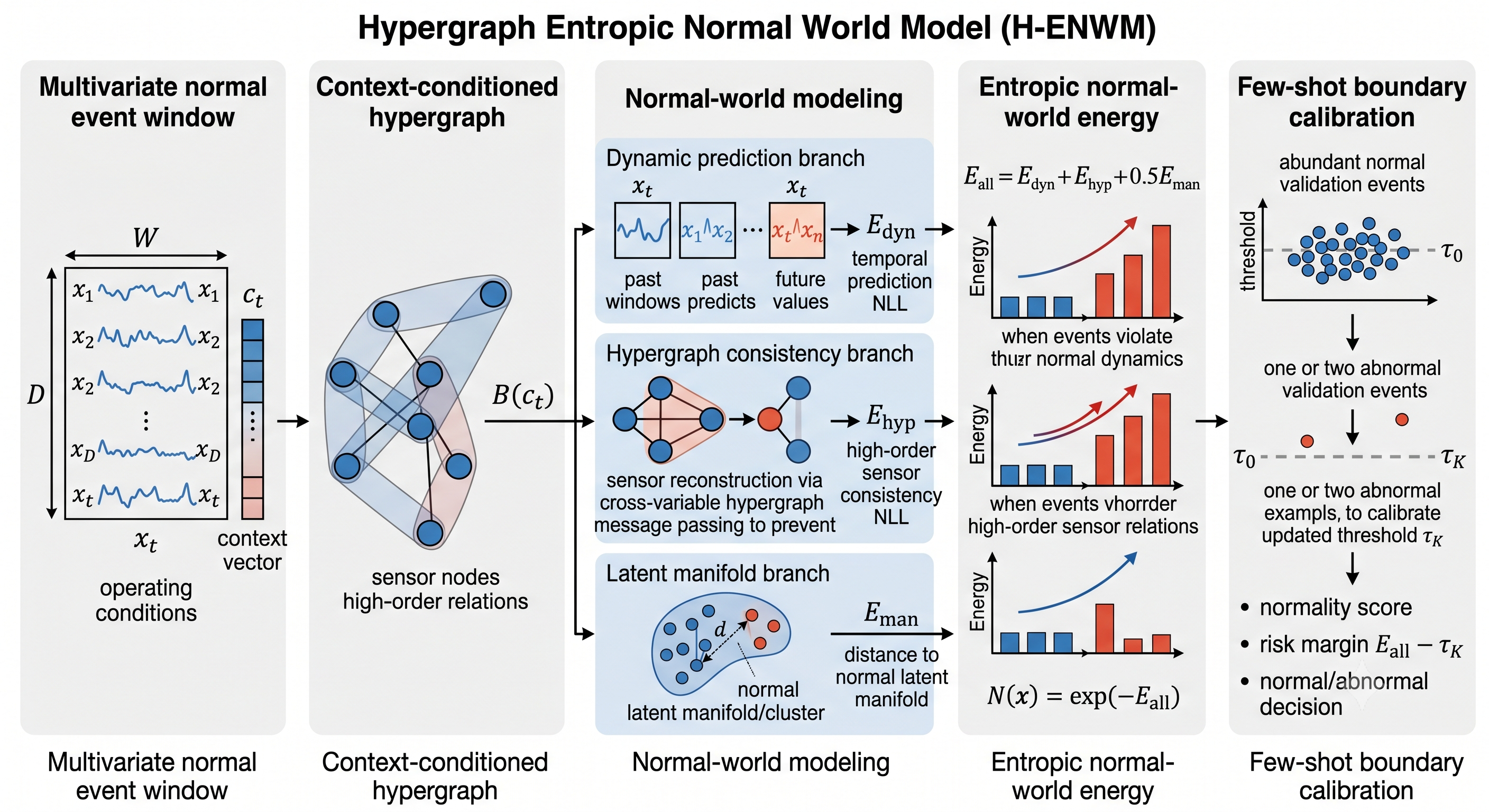}
}{
\fbox{\parbox{0.92\linewidth}{\centering
Place the workflow figure at \texttt{figures/workflow.png}.}}
}
\caption{Framework overview. A multivariate normal event window is represented as a context-conditioned hypergraph, processed by dynamic prediction, hypergraph consistency, and latent manifold branches, and scored by an entropic normal-world energy before few-shot boundary calibration.}
\label{fig:workflow}
\end{figure}

\subsection{Normal-World Learning Objective}

The normal world model is learned without abnormal labels.
For a normal event window \((x_t,c_t)\), the model should satisfy three requirements.
First, it should predict the future state under normal temporal dynamics.
Second, it should explain each sensor through its relation to other sensors.
Third, it should map normal events to a compact latent region.
We therefore learn a parameterized model \(M_\Theta\) by minimizing the expected normal-world surprise:
\begin{equation}
\min_\Theta
\mathbb{E}_{(x_t,c_t)\sim \mathcal{D}_N}
\left[
\Edyn(x_t,c_t)+\Ehyp(x_t,c_t)+\lambda_{\mathrm{man}}\Eman(x_t,c_t)
\right],
\end{equation}
where the three terms are defined below.
This objective does not require knowing which fault type an event belongs to.
It only requires examples of what the system looks like when it is operating normally.

\subsection{Why Hypergraphs for Multivariate Normal Worlds?}

Many multivariate systems are not collections of independent variables.
In a turbofan engine, for example, temperature, pressure, speed, and flow jointly describe a physical operating state.
The normal value of one sensor depends not only on another single sensor, but often on a group of sensors under a particular operating context.
An ordinary graph represents pairwise edges \(v_i\leftrightarrow v_j\).
This is useful but limited.
A pairwise edge can say that temperature and pressure are related.
It does not directly represent a high-order constraint such as ``temperature, pressure, speed, and fuel flow together form a valid thermal-power state.''

A hypergraph represents such group-wise relations directly.
We define a hypergraph \(\mathcal{G}_H=(\mathcal{V},\mathcal{E})\), where each variable is a node
\begin{equation}
\mathcal{V}=\{v_1,\ldots,v_D\},
\end{equation}
and each hyperedge \(e_m\subseteq\mathcal{V}\) denotes a high-order relation among a group of variables.
For instance, one hyperedge may connect temperature-like sensors, another may connect pressure and speed sensors, and another may connect variables whose joint state indicates the current operating regime.
This representation is important for the coupling-break validation.
If one sensor group is replaced by values from another normal window, every single variable can remain marginally normal.
The hyperedge-level relation, however, becomes inconsistent.

\begin{figure}[t]
\centering
\IfFileExists{figures/hypergraph_construction.png}{
\includegraphics[width=\linewidth]{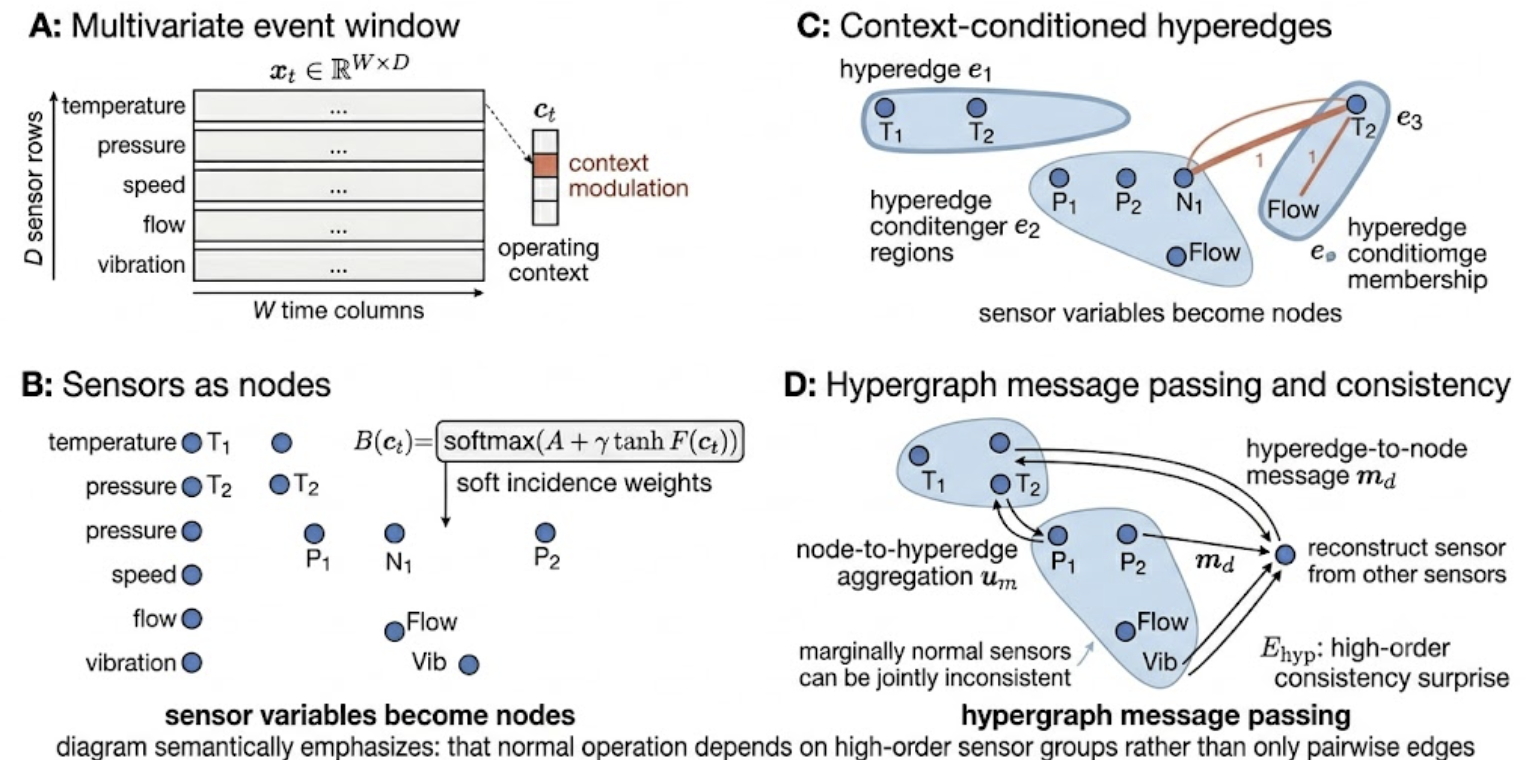}
}{
\fbox{\parbox{0.92\linewidth}{\centering
Place the hypergraph construction figure at \texttt{figures/hypergraph\_construction.png}.}}
}
\caption{Hypergraph construction. Sensors are represented as nodes, operating context modulates the incidence matrix, and hyperedges capture high-order sensor groups such as temperature-pressure-speed-flow relations.}
\label{fig:hypergraph-construction}
\end{figure}

\subsection{Context-Conditioned Hypergraph Construction}

The hypergraph incidence matrix is \(B(c_t)\in[0,1]^{D\times M}\), where \(B_{dm}(c_t)\) measures the membership strength of sensor \(d\) in hyperedge \(m\).
We use a soft incidence matrix rather than a hard binary one so that the model can learn overlapping sensor groups.
Unlike a fixed hypergraph, the incidence matrix is conditioned on operating context:
\begin{equation}
B(c_t)=\operatorname{softmax}_d
\left(
A + \gamma \tanh F_\eta(c_t)
\right),
\end{equation}
where \(A\) is a learnable base incidence-logit matrix, \(F_\eta(c_t)\) is a context-dependent adjustment, and \(\gamma\) controls contextual modulation.
The softmax is taken over sensors for each hyperedge so that a hyperedge forms a normalized group of variables.
When \(\gamma=0\), the model reduces to a context-invariant hypergraph.
When \(\gamma>0\), different operating conditions can induce different high-order relations.
This is essential for systems such as C-MAPSS FD002 and FD004, where the same sensor value may be normal in one operating regime but abnormal in another.

\subsection{Temporal Node Encoding and Hypergraph Message Passing}

For each sensor \(d\), the temporal segment is \(x_t^{(d)}\in\mathbb{R}^{W}\).
A temporal encoder maps each sensor segment to a node representation:
\begin{equation}
h_d=f_\theta^{\mathrm{temp}}(x_t^{(d)},c_t).
\end{equation}
The model aggregates node states into hyperedge states:
\begin{equation}
u_m=
\frac{
\sum_{d=1}^{D}B_{dm}(c_t)W_e h_d
}{
\sum_{d=1}^{D}B_{dm}(c_t)+\epsilon
}.
\end{equation}
Hyperedge messages are propagated back to nodes:
\begin{equation}
m_d=
\frac{
\sum_{m=1}^{M}B_{dm}(c_t)W_u u_m
}{
\sum_{m=1}^{M}B_{dm}(c_t)+\epsilon
}.
\end{equation}
The global latent state is \(z_t=\rho(\{h_d\}_{d=1}^D,\{u_m\}_{m=1}^M)\).

\subsection{Why Entropy Defines a Normality Score}

We now explain why information surprise and entropy provide a natural abnormality score.
Suppose a model learned from normal data assigns a conditional density \(q_\Theta(y|x,c)\) to a target observation \(y\).
The ideal code length for observing \(y\) under this normal model is \(-\log q_\Theta(y|x,c)\).
If \(y\) follows the normal world, this code length should be small.
If \(y\) violates normal dynamics or normal sensor relations, the code length should increase.
Thus, negative log-likelihood is not merely a training loss; it is an information-theoretic measure of how surprising an event is under the normal world.

For a Gaussian conditional distribution \(q_\Theta(y|x,c)=\mathcal{N}(\mu_\Theta,\Sigma_\Theta)\), the surprise decomposes into two parts:
\begin{equation}
-\log q_\Theta(y|x,c)
=
\frac{1}{2}(y-\mu_\Theta)^\top\Sigma_\Theta^{-1}(y-\mu_\Theta)
+
\frac{1}{2}\log|\Sigma_\Theta|
+ \mathrm{const}.
\end{equation}
The first term measures how far the observation is from the normal prediction.
The second term is proportional to the differential entropy of the Gaussian distribution:
\begin{equation}
H[\mathcal{N}(\mu_\Theta,\Sigma_\Theta)]
=
\frac{1}{2}\log\left((2\pi e)^k|\Sigma_\Theta|\right).
\end{equation}
Therefore, entropy enters the energy as a principled uncertainty penalty.
An event can be abnormal because it has a large residual, because the normal model is uncertain about how to explain it, or because both happen simultaneously.

This entropy-aware view is useful for abnormality detection because it produces a continuous normality score rather than only a binary decision.
The energy \(E_\Theta(x,c)\) can be interpreted as the amount of information needed to encode the event under the normal world model.
Low energy means the event is easy to explain by normal dynamics and relations.
High energy means the event requires extra information, which we interpret as departure from normality.

\subsection{Entropic Normal-World Energy}

Our full energy combines three forms of normal-world surprise.

\paragraph{Dynamic energy.}
Given \(z_t\), the model predicts the next sensor state:
\begin{equation}
p_\theta(x_{t+1}|x_t,c_t)
=
\mathcal{N}(\mu_\theta(z_t,c_t),\Sigma_\theta(z_t,c_t)).
\end{equation}
The dynamic energy is the Gaussian negative log-likelihood:
\begin{equation}
\Edyn
=
\frac{1}{2}r_t^\top\Sigma_\theta^{-1}r_t
+
\frac{1}{2}\log|\Sigma_\theta|,
\quad
r_t=x_{t+1}-\mu_\theta(z_t,c_t).
\end{equation}
This term combines prediction violation and predictive entropy.

\paragraph{Hypergraph consistency energy.}
Temporal prediction alone may miss relational inconsistency inside the current window.
Therefore, the hypergraph branch asks whether each sensor can be explained by the other sensors through high-order hypergraph messages.
To avoid a trivial self-reconstruction shortcut, the hypergraph head reconstructs each sensor using cross-variable hypergraph messages \(m_d\), not its own node encoding:
\begin{equation}
p_{\theta,d}(x_t^{(d)}|m_d,c_t)
=
\mathcal{N}(\mu_{\theta,d}(m_d),\sigma_{\theta,d}^2(m_d)).
\end{equation}
The hypergraph consistency energy is
\begin{equation}
\Ehyp
=
\frac{1}{D}
\sum_{d=1}^{D}
\left[
\frac{(x_t^{(d)}-\mu_{\theta,d})^2}{2\sigma_{\theta,d}^{2}}
+
\frac{1}{2}\log\sigma_{\theta,d}^{2}
\right].
\end{equation}
This term is high when a sensor value cannot be explained by its high-order relation to other sensors under the current context.

\paragraph{Latent manifold energy.}
The dynamic and hypergraph terms are local forms of surprise.
We also impose a global normal-region constraint in latent space.
Let \(\bar{z}_N\) be the center of latent states estimated from normal training windows.
Normal latent states should remain close to this learned normal center:
\begin{equation}
\Eman=\|z_t-\bar{z}_N\|_2^2.
\end{equation}
The full energy used for all external method comparisons is
\begin{equation}
\boxed{
\Eall
=
\Edyn+\Ehyp+0.5\Eman.
}
\label{eq:energy}
\end{equation}
Component energies are used only in ablation studies.

\subsection{Few-Shot Boundary Calibration}

After normal-world learning, the model parameters are frozen.
The remaining question is where the normal region should end.
In zero-shot mode, the boundary is estimated from normal validation windows:
\begin{equation}
\tau_0=
Q_{0.95}
\left(
\{ \Eall(x_i,c_i):(x_i,c_i)\in\mathcal{D}_{N}^{\mathrm{val}}\}
\right).
\end{equation}
Given \(K\) abnormal examples, only the threshold is calibrated:
\begin{equation}
\tau_K=\arg\max_{\tau} \operatorname{BalancedAccuracy}
\left(
\mathcal{D}_{N}^{\mathrm{val}},
\mathcal{D}_{A}^{K};
\tau
\right).
\end{equation}
The calibrated decision is \(\hat{y}=\mathbb{I}[\Eall(x,c)>\tau_K]\), and the risk margin is \(\Eall(x,c)-\tau_K\).
This calibration tests whether the normal-world model has already absorbed most of the knowledge needed for abnormality detection.

\section{Experiments}

\subsection{Verification Dataset: NASA C-MAPSS}

We use the NASA C-MAPSS turbofan engine degradation benchmark \citep{saxena2008cmapss} as the first verification dataset.
C-MAPSS is a simulated run-to-failure dataset for aircraft engines.
Each engine trajectory starts from a healthy state and evolves toward failure, and each time step contains operating settings and multivariate sensor measurements.
The benchmark is widely used in remaining useful life estimation and fault detection because it provides controlled degradation trajectories with known operating conditions and fault-mode settings.

This experiment should be interpreted as a \emph{verification} experiment rather than a final exhaustive benchmark.
At this stage, our goal is verification rather than exhaustive benchmarking.
We test whether the proposed normal-world energy can be learned quickly, evaluated reproducibly, and compared with representative anomaly-detection baselines in a controlled multivariate setting.
C-MAPSS fits this goal.
It is small enough for rapid iteration, provides complete run-to-failure trajectories, and contains both simple and complex regimes.
In the formal version of this work, we will extend the evaluation to additional industrial and clinical datasets.

C-MAPSS contains four subsets.
FD001 has one operating condition and one fault mode.
FD002 has six operating conditions and one fault mode.
FD003 has one operating condition and two fault modes.
FD004 has six operating conditions and two fault modes, making it the most challenging subset in this verification study.
This structure is important for our hypothesis.
A method that only models marginal sensor distributions may work in a single-condition subset.
It can fail when the same sensor value has different meanings under different operating contexts.
FD002 and FD004 therefore test whether context-conditioned normal-world modeling is useful.

\subsection{Experimental Protocol}

For each subset, engines are split into fitting, calibration, and evaluation groups.
The model is trained only on early healthy windows with \(\RUL\geq 125\).
Abnormal evaluation windows are defined by \(\RUL\leq 30\), while intermediate windows are used only for degradation-ordering analysis.
This protocol matches the intended use case: abundant normal operation is available, but abnormal examples are scarce or absent during model training.

We evaluate four calibration regimes.
In the zero-shot regime, the threshold is estimated only from normal validation windows.
In the \(K\)-shot regimes, \(K\in\{1,2,4,8\}\) abnormal calibration windows are used only to adjust the decision threshold; model weights are not updated.
The full-boundary setting uses all abnormal calibration windows and is reported as an upper-bound boundary calibration result.
For all comparisons with external methods, our method uses only the full energy \(\Eall\) in Eq.~\ref{eq:energy}; component energies are reserved for ablation studies.

We report AUROC, AUPRC, accuracy, precision, recall, F1, balanced accuracy, and Spearman correlation between energy and degradation level \(-\RUL\).
AUROC and AUPRC measure threshold-free ranking quality.
F1 and balanced accuracy measure thresholded detection quality after calibration.
The Spearman correlation measures whether energy increases monotonically as the engine approaches failure, which is useful for interpreting energy as a graded risk score rather than only a classifier output.

\subsection{Baselines}

We compare against both classical normal-data baselines and more recent neural time-series baselines.
The classical group includes four methods.
Gaussian last-state energy is based on Mahalanobis distance \citep{mahalanobis1936distance}.
We also include PCA reconstruction \citep{jolliffe2002principal}, ridge temporal prediction \citep{hoerl1970ridge}, and KNN distance \citep{cover1967nearest}.
These methods are important because they represent strong, simple normality estimators.
In particular, nearest-neighbor and Gaussian distances can be highly competitive when degradation appears as a direct state-space shift.

The neural group includes GRU forecasting \citep{cho2014learning}, Transformer forecasting \citep{vaswani2017attention}, TCN forecasting \citep{bai2018empirical}, and GDN-lite \citep{deng2021gdn}.
GRU, Transformer, and TCN baselines test whether standard temporal prediction errors are sufficient for this problem.
GDN-lite tests a graph-based relational model, which is especially relevant because our method argues for high-order hypergraph relations rather than only pairwise graph relations.
All baselines use the same train, calibration, and evaluation split as our method.
The main comparison reports every baseline by name.
We do not hide methods inside a best-of group.

\begin{table}[t]
\centering
\scriptsize
\resizebox{\linewidth}{!}{
\begin{tabular}{llllrrrrrrr}
\toprule
Subset & Group & Method & Protocol & AUROC & AUPRC & Acc & Prec & Recall & F1 & BAcc \\
\midrule
FD001 & Traditional & Gaussian last & Full & \textbf{0.9995} & \textbf{0.9990} & \textbf{0.9911} & 0.9850 & 0.9850 & \textbf{0.9850} & \textbf{0.9894} \\
FD001 & Traditional & PCA last-8 & Full & 0.6823 & 0.5946 & 0.6644 & 0.4465 & 0.5700 & 0.5007 & 0.6370 \\
FD001 & Traditional & Ridge predictor & Full & 0.8725 & 0.7993 & 0.8248 & 0.6961 & 0.7217 & 0.7087 & 0.7948 \\
FD001 & Traditional & KNN last-5 & Full & 0.9956 & 0.9928 & 0.9833 & \textbf{0.9863} & 0.9567 & 0.9712 & 0.9755 \\
FD001 & Deep & GRU forecast & Full & 0.9882 & 0.9755 & 0.9744 & 0.9676 & 0.9450 & 0.9562 & 0.9659 \\
FD001 & Deep & Transformer forecast & Full & 0.9811 & 0.9272 & 0.9626 & 0.9367 & 0.9367 & 0.9367 & 0.9551 \\
FD001 & Deep & TCN forecast & Full & 0.9877 & 0.9773 & 0.9705 & 0.9426 & 0.9583 & 0.9504 & 0.9669 \\
FD001 & Deep & GDN-lite & Full & 0.9612 & 0.9325 & 0.9011 & 0.8265 & 0.8417 & 0.8340 & 0.8838 \\
FD001 & Ours & NWM & 0-shot & 0.9934 & 0.9875 & 0.9596 & 0.8877 & \textbf{0.9883} & 0.9353 & 0.9680 \\
FD001 & Ours & NWM & 1-shot & 0.9949 & 0.9901 & 0.8888 & \textbf{1.0000} & 0.6233 & 0.7680 & 0.8117 \\
FD001 & Ours & NWM & 2-shot & 0.9949 & 0.9901 & 0.7825 & \textbf{1.0000} & 0.2633 & 0.4169 & 0.6317 \\
FD001 & Ours & NWM & Full & 0.9960 & 0.9885 & 0.9833 & 0.9579 & 0.9867 & 0.9721 & 0.9843 \\
\midrule
FD002 & Traditional & Gaussian last & Full & 0.5243 & 0.4649 & 0.4497 & 0.3972 & \textbf{0.9929} & 0.5674 & 0.5662 \\
FD002 & Traditional & PCA last-8 & Full & 0.9231 & 0.9030 & 0.8595 & 0.8266 & 0.7763 & 0.8007 & 0.8417 \\
FD002 & Traditional & Ridge predictor & Full & 0.5537 & 0.4258 & 0.5529 & 0.4123 & 0.5410 & 0.4680 & 0.5503 \\
FD002 & Traditional & KNN last-5 & Full & \textbf{0.9913} & \textbf{0.9895} & \textbf{0.9658} & 0.9803 & 0.9244 & \textbf{0.9515} & \textbf{0.9569} \\
FD002 & Deep & GRU forecast & Full & 0.5350 & 0.3801 & 0.4732 & 0.3853 & 0.7545 & 0.5101 & 0.5335 \\
FD002 & Deep & Transformer forecast & Full & 0.5315 & 0.3821 & 0.4634 & 0.3846 & 0.7936 & 0.5181 & 0.5342 \\
FD002 & Deep & TCN forecast & Full & 0.5420 & 0.3840 & 0.4793 & 0.3891 & 0.7590 & 0.5144 & 0.5393 \\
FD002 & Deep & GDN-lite & Full & 0.7417 & 0.7278 & 0.7565 & 0.7151 & 0.5487 & 0.6210 & 0.7120 \\
FD002 & Ours & NWM & 0-shot & 0.9173 & 0.8802 & 0.8052 & 0.8795 & 0.5378 & 0.6675 & 0.7479 \\
FD002 & Ours & NWM & 1-shot & 0.9384 & 0.9043 & 0.7656 & \textbf{1.0000} & 0.3551 & 0.5241 & 0.6776 \\
FD002 & Ours & NWM & 2-shot & 0.9384 & 0.9043 & 0.7202 & \textbf{1.0000} & 0.2301 & 0.3742 & 0.6151 \\
FD002 & Ours & NWM & Full & 0.9650 & 0.9486 & 0.8844 & 0.7765 & 0.9577 & 0.8576 & 0.9001 \\
\midrule
FD003 & Traditional & Gaussian last & Full & 0.9899 & 0.9615 & 0.9773 & 0.9224 & \textbf{0.9900} & 0.9550 & 0.9816 \\
FD003 & Traditional & PCA last-8 & Full & 0.8743 & 0.7780 & 0.8078 & 0.5786 & 0.7667 & 0.6595 & 0.7939 \\
FD003 & Traditional & Ridge predictor & Full & 0.8822 & 0.6997 & 0.8220 & 0.5971 & 0.8200 & 0.6910 & 0.8213 \\
FD003 & Traditional & KNN last-5 & Full & \textbf{0.9979} & \textbf{0.9956} & \textbf{0.9850} & 0.9577 & 0.9817 & \textbf{0.9695} & \textbf{0.9839} \\
FD003 & Deep & GRU forecast & Full & 0.9915 & 0.9837 & 0.9725 & 0.9555 & 0.9300 & 0.9426 & 0.9581 \\
FD003 & Deep & Transformer forecast & Full & 0.9963 & 0.9915 & 0.9802 & \textbf{0.9726} & 0.9450 & 0.9586 & 0.9682 \\
FD003 & Deep & TCN forecast & Full & 0.9717 & 0.9488 & 0.9369 & 0.8491 & 0.9000 & 0.8738 & 0.9244 \\
FD003 & Deep & GDN-lite & Full & 0.9913 & 0.9676 & 0.9454 & 0.8289 & 0.9767 & 0.8967 & 0.9560 \\
FD003 & Ours & NWM & 0-shot & 0.9894 & 0.9714 & 0.9409 & 0.8179 & 0.9733 & 0.8889 & 0.9519 \\
FD003 & Ours & NWM & 1-shot & 0.9869 & 0.9527 & 0.9037 & 0.8868 & 0.6917 & 0.7772 & 0.8317 \\
FD003 & Ours & NWM & 2-shot & 0.9869 & 0.9527 & 0.9571 & 0.9023 & 0.9233 & 0.9127 & 0.9456 \\
FD003 & Ours & NWM & Full & 0.9869 & 0.9527 & 0.9624 & 0.8733 & 0.9883 & 0.9273 & 0.9712 \\
\midrule
FD004 & Traditional & Gaussian last & Full & 0.4972 & 0.3444 & 0.6207 & 0.3087 & 0.4207 & 0.3561 & 0.5539 \\
FD004 & Traditional & PCA last-8 & Full & 0.8489 & 0.7590 & 0.8072 & 0.5940 & 0.7160 & 0.6493 & 0.7767 \\
FD004 & Traditional & Ridge predictor & Full & 0.5262 & 0.3296 & 0.5849 & 0.2926 & 0.4687 & 0.3602 & 0.5461 \\
FD004 & Traditional & KNN last-5 & Full & 0.9955 & 0.9906 & 0.9716 & 0.9219 & 0.9680 & 0.9444 & 0.9704 \\
FD004 & Deep & GRU forecast & Full & 0.5383 & 0.2606 & 0.5304 & 0.2710 & 0.5227 & 0.3569 & 0.5278 \\
FD004 & Deep & Transformer forecast & Full & 0.5010 & 0.2977 & 0.4069 & 0.2735 & 0.8327 & 0.4118 & 0.5491 \\
FD004 & Deep & TCN forecast & Full & 0.5316 & 0.2600 & 0.4576 & 0.2662 & 0.6693 & 0.3810 & 0.5283 \\
FD004 & Deep & GDN-lite & Full & 0.9796 & 0.9597 & 0.9405 & 0.8605 & 0.9087 & 0.8839 & 0.9299 \\
FD004 & Ours & NWM & 0-shot & 0.9843 & 0.9694 & 0.9566 & 0.9133 & 0.9127 & 0.9130 & 0.9419 \\
FD004 & Ours & NWM & 1-shot & 0.9748 & 0.9524 & 0.8597 & \textbf{1.0000} & 0.4373 & 0.6085 & 0.7187 \\
FD004 & Ours & NWM & 2-shot & 0.9748 & 0.9524 & 0.8565 & \textbf{1.0000} & 0.4247 & 0.5962 & 0.7123 \\
FD004 & Ours & NWM & Full & \textbf{0.9983} & \textbf{0.9950} & \textbf{0.9845} & 0.9613 & \textbf{0.9773} & \textbf{0.9693} & \textbf{0.9821} \\
\bottomrule
\end{tabular}
}
\caption{Main comparison on C-MAPSS. Every baseline is reported by method name under the same split. Our method is reported only with the full energy \( \Eall \). Best values within each subset are bolded.}
\label{tab:main}
\end{table}

\subsection{Experiment 1: Comparison with Representative Baselines}

Table~\ref{tab:main} summarizes the main results.
On FD001 and FD003, the Gaussian and KNN baselines are very strong.
They reach AUROC 0.9995 and 0.9979, respectively.
This is expected because both subsets contain a single operating condition.
When the context is fixed, degradation can look like a simple displacement in sensor space.
Distance-based methods can then provide a strong statistic.
Our zero-shot model remains close to these baselines, with AUROC 0.9934 on FD001 and 0.9894 on FD003.
No abnormal sample is used to learn its representation.

The value of normal-world modeling is clearer in the more complex subsets.
FD002 and FD004 contain multiple operating conditions.
A sensor state cannot be judged by its marginal value alone.
On FD002, the zero-shot model reaches AUROC 0.9173 and improves to 0.9650 after full-boundary calibration.
KNN remains strongest on FD002, but the gap between zero-shot and calibrated performance is informative.
The learned energy already contains useful normal-world structure, and a small amount of abnormal calibration improves the operating threshold.
On FD004, the most complex setting, our method gives the best overall result.
Zero-shot AUROC is 0.9843, already higher than GDN-lite at 0.9796.
Full-boundary calibration reaches \textbf{0.9983} AUROC, \textbf{0.9950} AUPRC, \textbf{0.9693} F1, and \textbf{0.9821} balanced accuracy.

\begin{figure}[t]
\centering
\IfFileExists{figures/fewshot_sota_comparison.png}{
\includegraphics[width=0.96\linewidth]{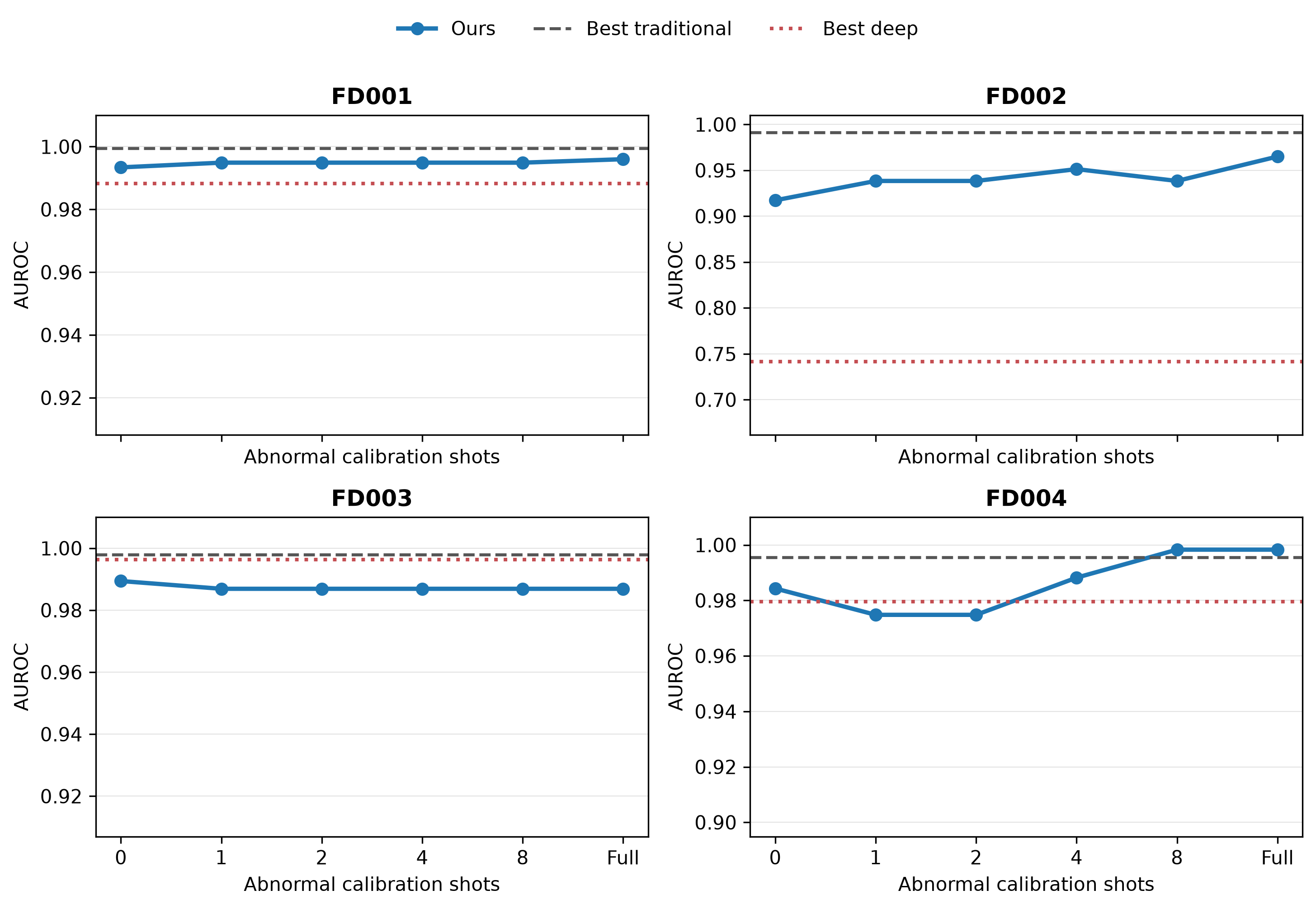}
}{
\fbox{\parbox{0.92\linewidth}{\centering
Place the few-shot comparison figure at \texttt{figures/fewshot\_sota\_comparison.png}.}}
}
\caption{Few-shot boundary calibration compared with the strongest traditional and neural baselines on each C-MAPSS subset. The model weights are fixed for all shots; abnormal examples only calibrate the threshold.}
\label{fig:fewshot-sota}
\end{figure}

Figure~\ref{fig:fewshot-sota} shows the effect of abnormal calibration shots.
The key observation is that zero-shot normal-world energy is already close to strong full-boundary baselines on most subsets.
This supports the central claim of the paper.
Most useful knowledge is learned from abundant normal events.
Abnormal examples mainly decide where to place the boundary.
More shots do not always improve thresholded metrics monotonically, because a tiny calibration set can be biased toward one engine or degradation stage.
The trend is still clear in the difficult multi-condition settings.
FD002 improves from 0.9173 to 0.9650 AUROC after full-boundary calibration, and FD004 improves from 0.9843 to 0.9983.

These results also show why standard forecasting error is not enough.
The GRU, Transformer, and TCN baselines predict future values, but their performance drops sharply on FD002 and FD004.
Temporal prediction alone is weak when context and sensor coupling change across regimes.
GDN-lite is stronger because it models relations, but it still relies on pairwise graph edges.
Our method adds context-conditioned high-order hypergraph consistency and a normal-manifold term.
This combination helps most on FD004, where abnormality can be a violation of a high-order relation rather than a large shift in one sensor.

Table~\ref{tab:main-fd004} expands the FD004 comparison because this subset is the closest setting to our target use case.
It contains multiple operating conditions and multiple fault modes.
The zero-shot normal-world energy already outperforms the lightweight deep baselines in AUROC, AUPRC, F1, and balanced accuracy.
After full-boundary calibration, it also exceeds the strongest KNN baseline on the same split.

\begin{table}[H]
\centering
\scriptsize
\resizebox{\linewidth}{!}{
\begin{tabular}{lllrrrrrrrr}
\toprule
Group & Method & Protocol & AUROC & AUPRC & Acc & Prec & Recall & F1 & BAcc & rho \\
\midrule
Traditional & Gaussian last & Full boundary & 0.4972 & 0.3444 & 0.6207 & 0.3087 & 0.4207 & 0.3561 & 0.5539 & -0.0002 \\
Traditional & PCA last-8 & Full boundary & 0.8489 & 0.7590 & 0.8072 & 0.5940 & 0.7160 & 0.6493 & 0.7767 & 0.3963 \\
Traditional & Ridge predictor & Full boundary & 0.5262 & 0.3296 & 0.5849 & 0.2926 & 0.4687 & 0.3602 & 0.5461 & 0.0331 \\
Traditional & KNN last-5 & Full boundary & 0.9955 & 0.9906 & 0.9716 & 0.9219 & 0.9680 & 0.9444 & 0.9704 & 0.6634 \\
\midrule
Deep & GRU forecast & Full boundary & 0.5383 & 0.2606 & 0.5304 & 0.2710 & 0.5227 & 0.3569 & 0.5278 & 0.0594 \\
Deep & Transformer forecast & Full boundary & 0.5010 & 0.2977 & 0.4069 & 0.2735 & 0.8327 & 0.4118 & 0.5491 & 0.0136 \\
Deep & TCN forecast & Full boundary & 0.5316 & 0.2600 & 0.4576 & 0.2662 & 0.6693 & 0.3810 & 0.5283 & 0.0611 \\
Deep & GDN-lite & Full boundary & 0.9796 & 0.9597 & 0.9405 & 0.8605 & 0.9087 & 0.8839 & 0.9299 & 0.6106 \\
\midrule
Ours & Ours & 0-shot & 0.9843 & 0.9694 & 0.9566 & 0.9133 & 0.9127 & 0.9130 & 0.9419 & 0.6053 \\
Ours & Ours & 1-shot & 0.9748 & 0.9524 & 0.8597 & 1.0000 & 0.4373 & 0.6085 & 0.7187 & 0.5754 \\
Ours & Ours & 2-shot & 0.9748 & 0.9524 & 0.8565 & 1.0000 & 0.4247 & 0.5962 & 0.7123 & 0.5754 \\
Ours & Ours & 4-shot & 0.9882 & 0.9764 & 0.9214 & 0.9961 & 0.6873 & 0.8134 & 0.8432 & 0.6285 \\
Ours & Ours & 8-shot & 0.9983 & 0.9950 & 0.9177 & 0.9951 & 0.6733 & 0.8032 & 0.8361 & 0.7293 \\
Ours & Ours & Full boundary & 0.9983 & 0.9950 & 0.9845 & 0.9613 & 0.9773 & 0.9693 & 0.9821 & 0.7293 \\
\bottomrule
\end{tabular}
}
\caption{Detailed comparison on C-MAPSS FD004. All methods use the same train/calibration/evaluation split.}
\label{tab:main-fd004}
\end{table}

\subsection{Experiment 2: Component and Energy Ablation}

We next ablate the three energy components in Eq.~\ref{eq:energy}.
This experiment asks which part of the normal-world energy is responsible for detection and whether the proposed energy decomposition is meaningful.
We evaluate single-component energies, pairwise combinations, the main full energy, and several alternative weighting schemes.
All ablation results use the same zero-shot boundary protocol so that the comparison reflects the learned energy itself rather than the amount of abnormal calibration.

\begin{table}[t]
\centering
\scriptsize
\resizebox{\linewidth}{!}{
\begin{tabular}{lrrrrrr}
\toprule
Energy variant & Avg. AUROC & Avg. AUPRC & Avg. BAcc & Avg. F1 & Avg. \(\rho\) & FD004 AUROC \\
\midrule
\(\Edyn\) only & 0.7529 & 0.6759 & 0.7433 & 0.5573 & 0.3025 & 0.5130 \\
\(\Ehyp\) only & 0.9803 & 0.9665 & \textbf{0.9297} & \textbf{0.8860} & 0.6385 & \textbf{0.9983} \\
\(\Eman\) only & 0.7475 & 0.6456 & 0.7350 & 0.5127 & 0.3614 & 0.4981 \\
\(\Edyn+\Ehyp\) & 0.9739 & 0.9586 & 0.9170 & 0.8741 & 0.5951 & 0.9882 \\
\(\Edyn+\Eman\) & 0.7526 & 0.6548 & 0.7327 & 0.5028 & 0.3542 & 0.4949 \\
\(\Ehyp+\Eman\) & 0.9787 & 0.9608 & 0.9190 & 0.8718 & \textbf{0.6578} & 0.9960 \\
All, equal weights & 0.9640 & 0.9389 & 0.8881 & 0.8303 & 0.6181 & 0.9748 \\
All, main weights & 0.9711 & 0.9521 & 0.9024 & 0.8512 & 0.6167 & 0.9843 \\
Dynamic-dominant & 0.9102 & 0.8538 & 0.8181 & 0.7162 & 0.5184 & 0.8657 \\
Hypergraph-dominant & \textbf{0.9813} & \textbf{0.9666} & 0.9268 & 0.8830 & 0.6500 & 0.9974 \\
Manifold-dominant & 0.9154 & 0.8442 & 0.8067 & 0.6881 & 0.5707 & 0.8679 \\
\bottomrule
\end{tabular}
}
\caption{Component and energy-weight ablation. Average metrics are computed over FD001--FD004 under zero-shot boundary calibration. FD004 AUROC is reported separately because it is the most complex multi-condition, multi-fault subset.}
\label{tab:ablation}
\end{table}

Table~\ref{tab:ablation} shows that the hypergraph consistency energy is the dominant component for abnormality detection.
Using \(\Ehyp\) alone achieves 0.9803 average AUROC and 0.9983 AUROC on FD004.
This supports the central design choice of the method: in multivariate systems, many abnormal events are not single-sensor outliers but violations of high-order sensor consistency.
The result is especially clear on FD004, where temporal dynamics alone is close to random ranking with AUROC 0.5130, while the hypergraph energy separates abnormal windows almost perfectly.

The dynamic and manifold terms are nevertheless not redundant.
The dynamic energy captures whether an event follows normal temporal evolution, and it performs well on FD003, where degradation is more temporally ordered under a single operating condition.
The manifold energy captures global departure from the latent normal region and improves degradation ordering in several settings.
For example, \(\Ehyp+\Eman\) gives the best average Spearman correlation, 0.6578.
This indicates that the manifold term helps energy behave like a graded degradation score rather than only a binary detector.
The three terms therefore have different roles.
\(\Ehyp\) is the strongest detector of high-order inconsistency.
\(\Edyn\) contributes temporal evolution information.
\(\Eman\) regularizes the global normal region.

The weight ablation further shows why the energy should not be treated as a purely arbitrary sum.
Hypergraph-dominant weighting gives the best average AUROC and AUPRC, while the main all-energy weighting remains a stable unified choice used for external comparisons.
In the current verification experiment, the strongest zero-shot ranking is often obtained by emphasizing \(\Ehyp\), especially in multi-condition subsets.
This observation is consistent with the motivation of the paper: the most informative signal for normal-world violation is often the breakdown of context-dependent high-order relations.

\section{Mechanistic Validation: Does the Energy Encode World Knowledge?}

High AUROC does not by itself prove that a model has learned a normal world.
A detector can obtain a strong score by mapping windows directly to labels.
It can also exploit simple marginal deviations in individual sensors.
Our claim is stronger.
The model should learn which unseen healthy states are valid, how degradation moves away from normality, how sensors should be coupled, and how context changes the meaning of a sensor configuration.
We therefore design mechanistic validation tests.
Each test probes a different world-model property.
Each test is also hard for a model that only learns a superficial input-output mapping.

\subsection{Unseen Normal Acceptance}

The first requirement of a normal world model is that it should not simply memorize the training engines.
If the model only learns a narrow mapping from familiar trajectories to low energy, then healthy windows from unseen engines may be rejected as abnormal.
In contrast, a model that has learned the normal world should accept held-out healthy engines as long as their dynamics and sensor couplings remain compatible with normal operation.

We evaluate this property using held-out healthy windows from engines that are not used for fitting the model.
The zero-shot boundary \(\tau_0\) is estimated from normal validation windows, and we compute the fraction of unseen healthy windows whose energy remains below this boundary:
\begin{equation}
\operatorname{Accept}=
\frac{1}{|\mathcal{D}_{N}^{\mathrm{test}}|}
\sum_{(x,c)\in\mathcal{D}_{N}^{\mathrm{test}}}
\mathbb{I}[\Eall(x,c)\leq\tau_0].
\end{equation}
We also report the mean and median risk margin \(\Eall(x,c)-\tau_0\), where negative values indicate that an unseen healthy window lies safely inside the learned normal region.

\begin{table}[t]
\centering
\scriptsize
\resizebox{\linewidth}{!}{
\begin{tabular}{lrrrr}
\toprule
Subset & Held-out normal windows & Acceptance rate & Mean margin & Median margin \\
\midrule
FD001 & 1432 & 0.9476 & -1.9677 & -2.2717 \\
FD002 & 2732 & 0.9579 & -4.1190 & -4.4925 \\
FD003 & 1872 & 0.9306 & -3.1286 & -4.0371 \\
FD004 & 4516 & 0.9712 & -2.4303 & -2.6814 \\
\bottomrule
\end{tabular}
}
\caption{Unseen normal acceptance. The learned energy accepts held-out healthy engines and assigns negative risk margins to normal windows not seen during fitting.}
\label{tab:normal-acceptance}
\end{table}

Table~\ref{tab:normal-acceptance} shows that the acceptance rate is above 0.93 on all four subsets and reaches 0.9712 on FD004.
The margins are also consistently negative, which means that most unseen normal windows are not merely close to the boundary but lie inside the normal region.
This supports the interpretation that the model has learned a reusable normal-state structure rather than memorizing engine-specific trajectories.

\subsection{Degradation Ordering}

The second probe asks whether energy behaves like a physical departure score.
A pure classifier mapping can separate late abnormal windows from early normal windows without learning the direction of degradation.
A normal world model should do more: as the engine gradually leaves the normal operating regime, the normal-world energy should increase monotonically.
This is important because the proposed score is intended to be a graded normality measure, not only a binary abnormal label.

We split trajectories into early, middle, and late degradation stages and compare their mean energy.
We also compute Spearman correlation between energy and degradation level \(-\RUL\).
The model is not trained with RUL regression labels in this test; the ordering is used only as an external physical validation of the learned energy.

\begin{table}[t]
\centering
\scriptsize
\resizebox{\linewidth}{!}{
\begin{tabular}{lrrrr}
\toprule
Subset & Early energy & Middle energy & Late energy & Spearman \(\rho(E,-\RUL)\) \\
\midrule
FD001 & -0.1513 & 3.8759 & 46.6377 & 0.6875 \\
FD002 & -0.0198 & 0.7195 & 5.7298 & 0.5118 \\
FD003 & 0.3738 & 34.6670 & 118.4837 & 0.6621 \\
FD004 & 0.0159 & 1.3191 & 8.6544 & 0.6053 \\
\bottomrule
\end{tabular}
}
\caption{Degradation ordering. Energy increases from early to middle to late degradation stages, even though the model is trained from normal windows rather than RUL labels.}
\label{tab:degradation-ordering}
\end{table}

Table~\ref{tab:degradation-ordering} shows a consistent early-to-middle-to-late increase on all subsets.
For example, FD004 energy rises from 0.0159 to 1.3191 to 8.6544, with Spearman correlation 0.6053.
This behavior is difficult to explain as a static mapping to abnormal labels because the model is never trained to rank RUL stages.
Instead, the energy surface appears to encode a meaningful direction away from the normal world.

\begin{figure}[t]
\centering
\begin{minipage}{0.48\linewidth}
\centering
\includegraphics[width=\linewidth]{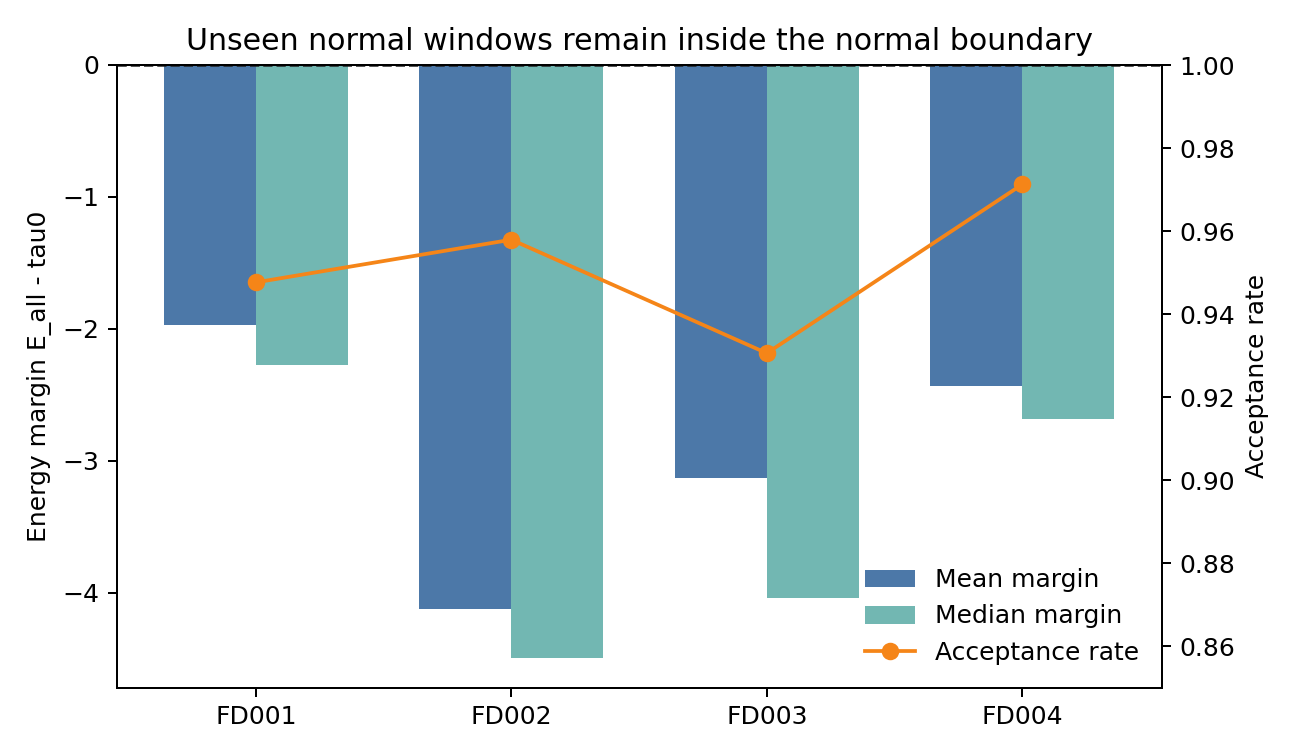}
\end{minipage}
\hfill
\begin{minipage}{0.48\linewidth}
\centering
\includegraphics[width=\linewidth]{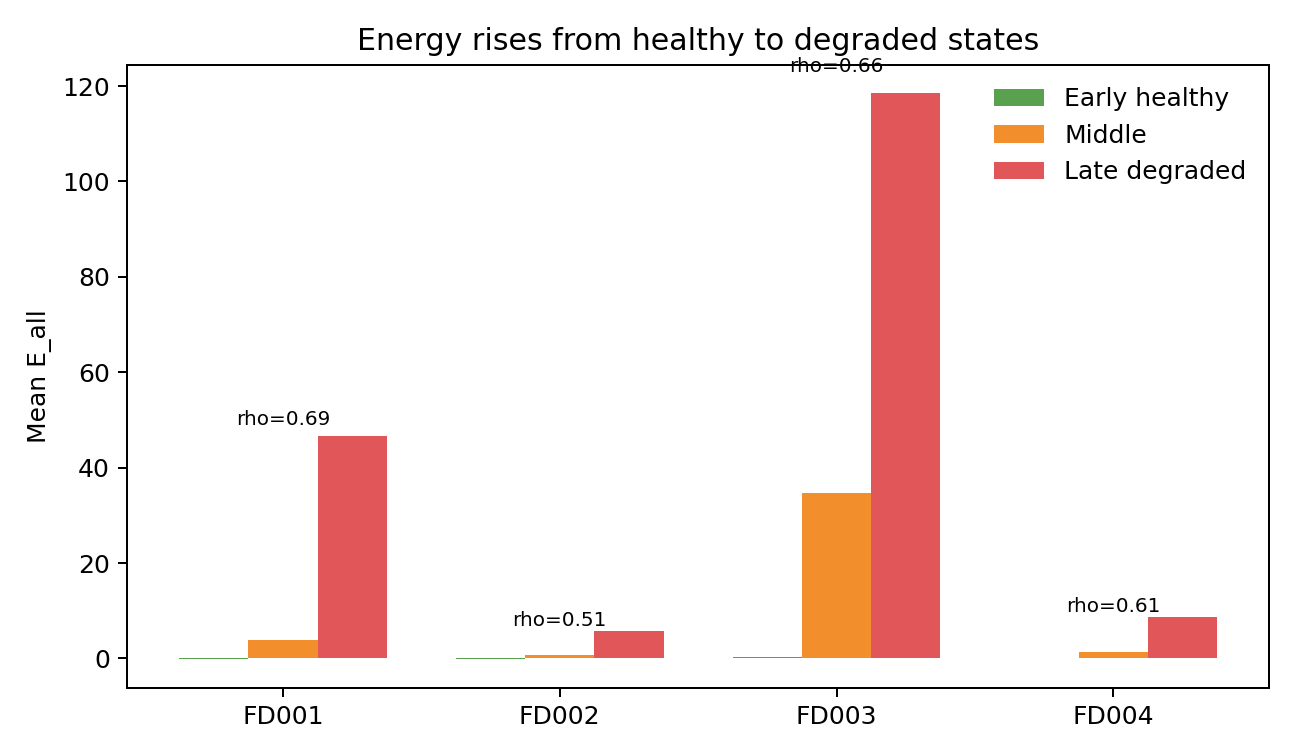}
\end{minipage}
\caption{Normal-world energy accepts held-out healthy engines and increases along degradation trajectories.}
\label{fig:normal-rul}
\end{figure}

\subsection{Cross-Variable Coupling Break}

The coupling-break test is the most direct test of whether the model learned world structure rather than marginal sensor statistics.
In a normal engine, temperature, pressure, speed, and flow do not vary independently.
They form a coupled physical state.
We therefore construct a counterfactual window in which every replaced sensor value is still drawn from a normal window, but the combination no longer necessarily corresponds to one coherent physical state.
Specifically, for a sensor group \(G\), we replace that group in a normal window with the same group from another normal window:
\begin{equation}
x'^{(G)}=\tilde{x}^{(G)},\quad x'^{(-G)}=x^{(-G)}.
\end{equation}
The corrupted window is deliberately subtle: each replaced sensor remains marginally plausible, so a model that only learns per-sensor distributions should assign it low energy.
A model that understands the normal world should detect that the joint high-order relation is broken.

We measure the energy increase and boundary crossing rate as
\begin{equation}
\Delta E = \Eall(x')-\Eall(x),
\qquad
\operatorname{BCR}=\Pr[\Eall(x')>\tau_0].
\end{equation}
We distinguish context-matched replacement, where the donor window comes from the same operating regime, from context-mismatched replacement, where the donor window comes from a different operating regime.
Context-matched replacement is a mechanism-preserving perturbation; context-mismatched replacement is a mechanism-breaking perturbation.

\begin{table}[t]
\centering
\scriptsize
\resizebox{\linewidth}{!}{
\begin{tabular}{lrrrr}
\toprule
Subset & Matched \(\Delta E\) & Mismatched \(\Delta E\) & Matched BCR & Mismatched BCR \\
\midrule
FD001 & 6.8050 & 11.3370 & 0.2998 & 0.2957 \\
FD002 & 0.3404 & 3010.6119 & 0.0545 & \textbf{1.0000} \\
FD003 & 28.5320 & 2.7532 & 0.2751 & 0.2376 \\
FD004 & 0.4220 & 3294.1372 & 0.0893 & \textbf{1.0000} \\
\bottomrule
\end{tabular}
}
\caption{Mechanism-preserving versus mechanism-breaking coupling perturbations. The context-mismatched coupling break is decisive on FD002 and FD004, the two multi-condition subsets.}
\label{tab:coupling-summary}
\end{table}

Table~\ref{tab:coupling-summary} shows that the effect is strongest exactly where it should be strongest: the multi-condition subsets FD002 and FD004.
On FD004, context-matched replacement changes energy only mildly on average, with mean \(\Delta E=0.4220\) and BCR 0.0893.
Context-mismatched replacement produces a very large mean \(\Delta E=3294.1372\) and BCR 1.0000.
This is evidence for context-dependent world knowledge.
The model is not simply asking whether a sensor value is normal in isolation.
It asks whether the full sensor group is coherent under the current operating condition.
FD001 and FD003 have a single operating condition, so the context-mismatch interpretation is less meaningful there, which is why the decisive separation appears in FD002 and FD004.

\begin{table}[t]
\centering
\scriptsize
\resizebox{\linewidth}{!}{
\begin{tabular}{lrrrr}
\toprule
FD004 sensor group & Matched \(\Delta E\) & Mismatched \(\Delta E\) & Matched BCR & Mismatched BCR \\
\midrule
Single sensor & 0.0361 & 1173.7075 & 0.0300 & \textbf{1.0000} \\
Temperature-like group & 0.6799 & 4063.6504 & 0.1415 & \textbf{1.0000} \\
Pressure/speed-like group & 0.6556 & 4311.6948 & 0.1160 & \textbf{1.0000} \\
Random five-sensor group & 0.3163 & 3627.4961 & 0.0695 & \textbf{1.0000} \\
\bottomrule
\end{tabular}
}
\caption{FD004 coupling-break details. Sensor groups sampled from a normal but context-mismatched window consistently violate the learned normal-world boundary.}
\label{tab:fd004-coupling-detail}
\end{table}

Table~\ref{tab:fd004-coupling-detail} further shows that the result is not caused by one special sensor group.
Single-sensor, temperature-like, pressure/speed-like, and random five-sensor replacements all have low boundary-crossing rates when context is matched, but all reach BCR 1.0000 when context is mismatched.
This is the clearest evidence that the energy captures high-order relational consistency.
The corrupted samples are built entirely from normal sensor values, so marginal normality is not enough to pass the test.

\begin{figure}[H]
\centering
\includegraphics[width=0.98\linewidth,height=0.78\textheight,keepaspectratio]{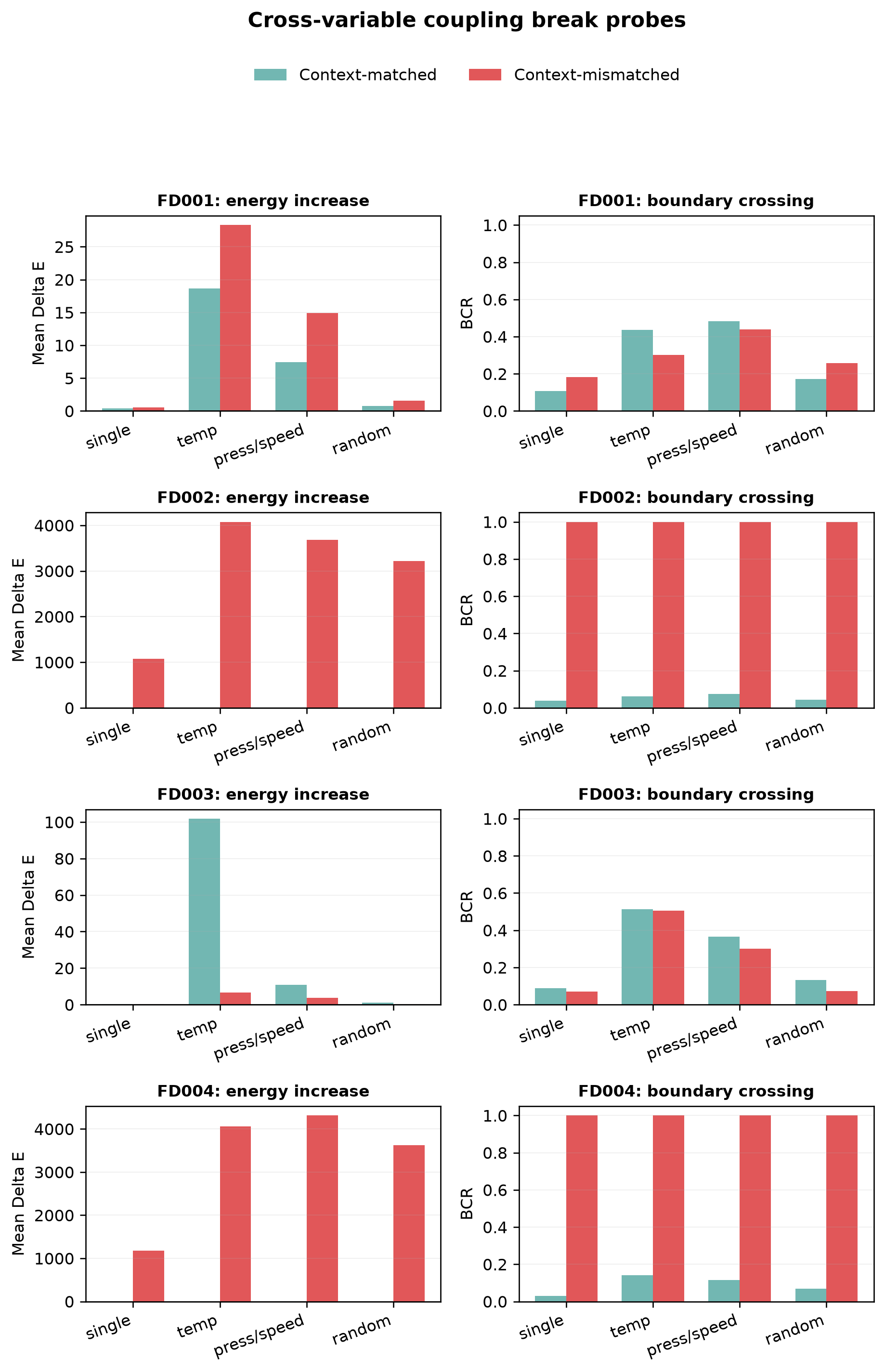}
\caption{Cross-variable coupling break. Each row corresponds to one C-MAPSS subset; the left column reports energy increase and the right column reports boundary crossing rate. Context-mismatched sensor replacements sharply increase normal-world energy and cross the zero-shot boundary in multi-condition subsets.}
\label{fig:coupling}
\end{figure}

\subsection{Context Counterfactual Consistency}

The previous test changes sensor values.
We also test the opposite counterfactual: keep the sensor window fixed but change the operating context.
This experiment is motivated by a key property of physical systems: the same sensor pattern can be normal or abnormal depending on the operating condition.
If the model ignores context or learns only a direct sensor-to-score mapping, changing the context should have little effect.
If it learns a context-conditioned normal world, then replacing the true context with an incompatible context should sharply increase energy.

For each normal window, we compare two counterfactuals.
The matched-context counterfactual uses a context sampled from the same operating regime, while the mismatched-context counterfactual uses a context sampled from a different regime.
We again measure \(\Delta E\) and BCR.

\begin{table}[t]
\centering
\scriptsize
\resizebox{\linewidth}{!}{
\begin{tabular}{lrrrr}
\toprule
Subset & Matched context \(\Delta E\) & Mismatched context \(\Delta E\) & Matched context BCR & Mismatched context BCR \\
\midrule
FD001 & 0.0001 & 0.0813 & 0.0524 & 0.0622 \\
FD002 & 0.0000 & 13036.5762 & 0.0390 & \textbf{1.0000} \\
FD003 & -0.0001 & 1.0875 & 0.0694 & 0.0855 \\
FD004 & 0.0000 & 14447.3408 & 0.0280 & \textbf{1.0000} \\
\bottomrule
\end{tabular}
}
\caption{Context counterfactual consistency. Changing context within the same regime has almost no effect, while mismatched context sharply increases energy on multi-condition subsets.}
\label{tab:context-counterfactual}
\end{table}

Table~\ref{tab:context-counterfactual} shows that matched-context changes have nearly zero mean effect.
This is important: it means the model is not hypersensitive to arbitrary context perturbations.
On FD002 and FD004, mismatched context causes extremely large energy increases and BCR 1.0000.
This directly supports the claim that the learned hypergraph is context-conditioned.
The model has learned not only what a normal sensor window looks like, but also which operating context makes that window physically plausible.

\subsection{Few-Shot Boundary Efficiency}

Finally, we test whether abnormal examples are needed to learn the representation or only to calibrate the boundary.
If the method were mainly learning an abnormal-label mapping, then zero-shot performance should be poor and many abnormal samples would be needed to shape the decision surface.
In our protocol, the model weights are fixed after normal-world learning.
The \(K\)-shot examples only move the scalar threshold \(\tau_K\).
Therefore, strong zero-shot performance and small gaps to the full-boundary result indicate that the useful information is already contained in the learned normal-world energy.

\begin{table}[t]
\centering
\scriptsize
\resizebox{\linewidth}{!}{
\begin{tabular}{lrrrrrr}
\toprule
Subset & 0-shot & 1-shot & 2-shot & 4-shot & 8-shot & Full boundary \\
\midrule
FD001 & 0.9934 & 0.9949 & 0.9949 & 0.9949 & 0.9949 & 0.9960 \\
FD002 & 0.9173 & 0.9384 & 0.9384 & 0.9513 & 0.9384 & 0.9650 \\
FD003 & 0.9894 & 0.9869 & 0.9869 & 0.9869 & 0.9869 & 0.9869 \\
FD004 & 0.9843 & 0.9748 & 0.9748 & 0.9882 & 0.9983 & 0.9983 \\
\bottomrule
\end{tabular}
}
\caption{Few-shot boundary efficiency measured by AUROC. Model parameters are frozen; abnormal examples only calibrate the threshold.}
\label{tab:fewshot-efficiency}
\end{table}

Table~\ref{tab:fewshot-efficiency} shows that zero-shot energy is already close to the full-boundary result on FD001, FD003, and FD004.
On FD004, zero-shot AUROC is 0.9843 and reaches 0.9983 with eight abnormal examples or full-boundary calibration.
On FD002, the gap is larger but still moderate, improving from 0.9173 to 0.9650.
These results support the intended learning mechanism: abnormal samples are not used to learn the world model; they mainly calibrate where the learned normal world should be cut by the decision boundary.

\begin{figure}[H]
\centering
\includegraphics[width=0.98\linewidth]{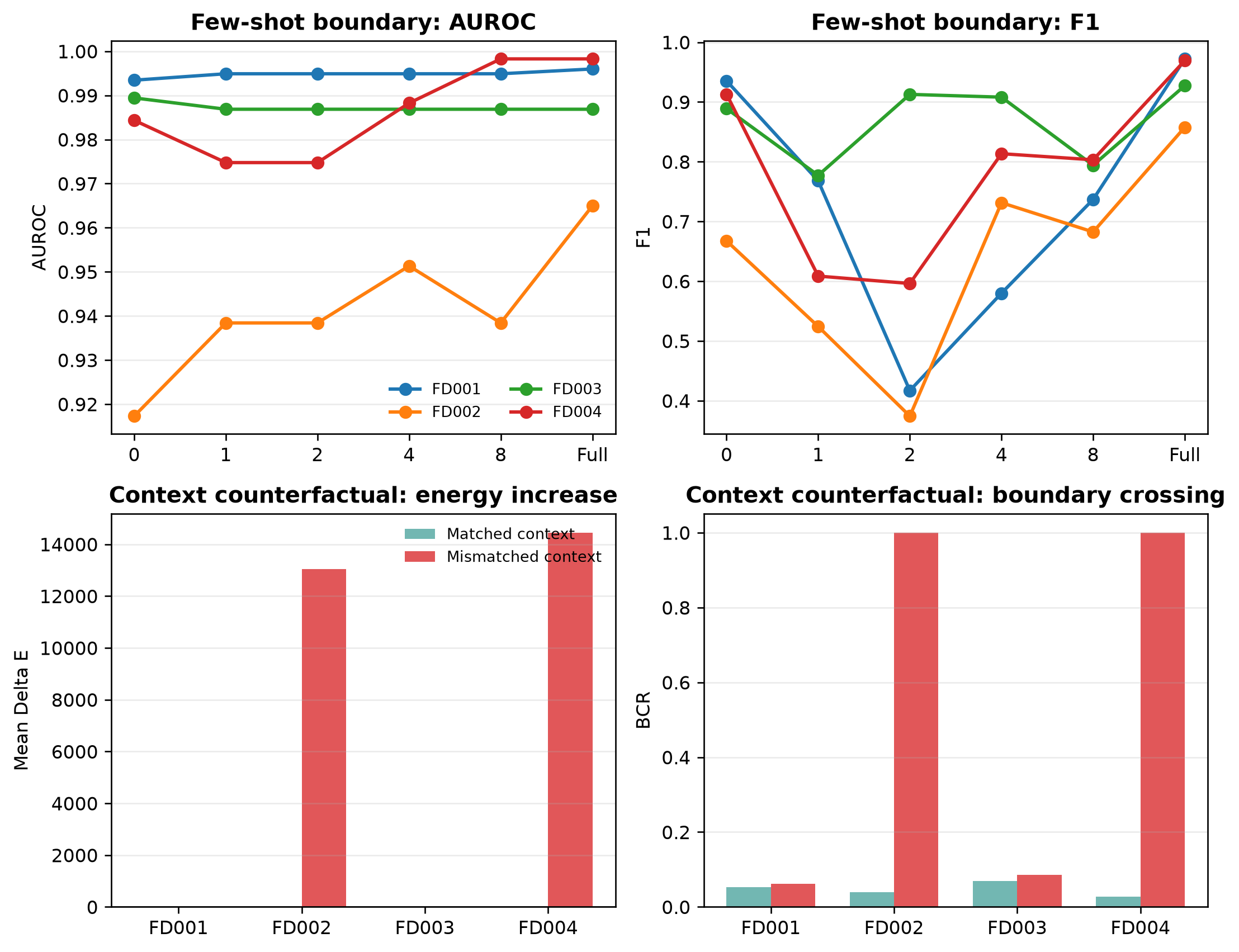}
\caption{Few-shot boundary calibration and context counterfactual consistency. The top row reports few-shot AUROC and F1 with fixed model weights, while the bottom row reports context-counterfactual energy increase and boundary crossing rate.}
\label{fig:fewshot}
\end{figure}

Together, these probes address the central concern of this paper.
Unseen normal acceptance tests generalization inside the normal world.
Degradation ordering tests whether energy has a physical direction.
Coupling break tests whether the model captures high-order sensor relations rather than marginal distributions.
Context counterfactuals test whether operating conditions are part of the learned world state.
Few-shot boundary efficiency tests whether abnormal examples calibrate the boundary instead of learning the representation.
The positive evidence across these probes supports our interpretation.
The learned energy is not only a direct mapping from windows to anomaly scores.
It encodes structured knowledge of the normal operating world.

\subsection{Hyperparameter Sensitivity}

We finally test whether the method depends on a narrow hyperparameter choice.
This is a sensitivity probe, not a new main comparison.
All runs use the two multi-condition subsets FD002 and FD004 with a shortened 60-epoch schedule.
We vary one hyperparameter at a time around the base setting \(W=30\), hidden dimension 128, 16 hyperedges, context modulation strength \(\gamma=0.25\), and threshold quantile 0.95.
The goal is to check whether the normal-world energy remains effective under reasonable settings.
The sweep also identifies which design choices are most sensitive.

\begin{table}[t]
\centering
\scriptsize
\resizebox{\linewidth}{!}{
\begin{tabular}{llllrrr}
\toprule
Hyperparameter & Tested values & Best avg. AUROC setting & Best avg. BAcc setting & Best avg. AUROC & Best avg. BAcc & F1 at best BAcc \\
\midrule
Window length \(W\) & 20, 30, 50 & 30 & 30 & 0.9561 & 0.8670 & 0.8310 \\
Hidden dimension & 64, 128, 256 & 128 & 128 & 0.9561 & 0.8670 & 0.8310 \\
Number of hyperedges \(M\) & 8, 16, 32 & 8 & 8 & \textbf{0.9807} & \textbf{0.9345} & \textbf{0.9112} \\
Context modulation \(\gamma\) & 0, 0.1, 0.25, 0.5, 1.0 & 0 & 0.25 & 0.9568 & 0.8670 & 0.8310 \\
Threshold quantile & 0.90, 0.95, 0.975 & tie & 0.90 & 0.9561 & 0.8898 & 0.8392 \\
\bottomrule
\end{tabular}
}
\caption{Hyperparameter sensitivity on FD002 and FD004. Metrics are averaged over the two multi-condition subsets. The sweep varies one hyperparameter at a time around the base setting.}
\label{tab:hyperparam-sensitivity}
\end{table}

\begin{figure}[t]
\centering
\includegraphics[width=0.98\linewidth]{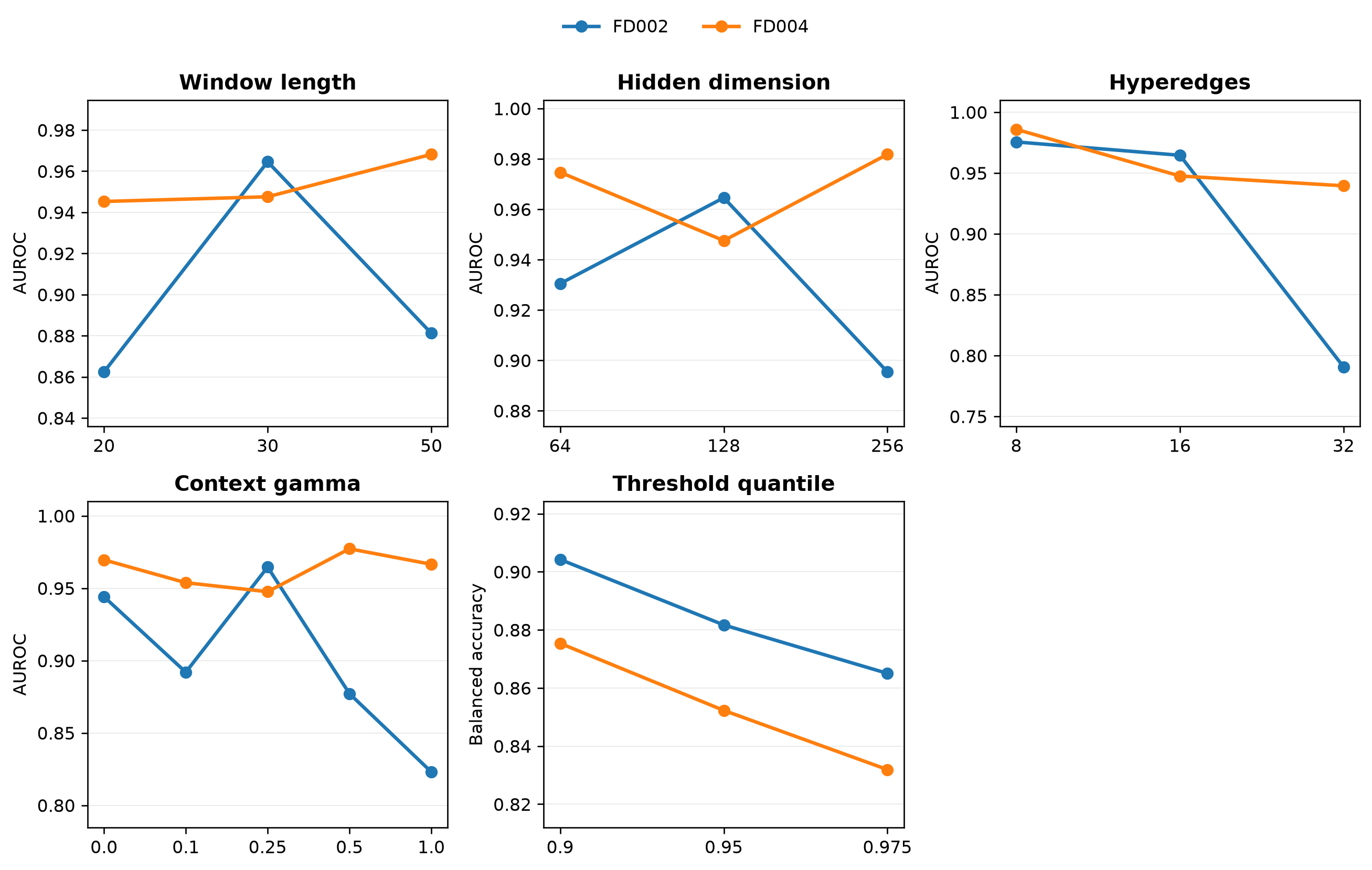}
\caption{Hyperparameter sensitivity curves on FD002 and FD004. AUROC is shown for model-structure hyperparameters, while balanced accuracy is shown for the threshold quantile because the quantile affects the operating point rather than ranking.}
\label{fig:hyperparam-sensitivity}
\end{figure}

Table~\ref{tab:hyperparam-sensitivity} and Figure~\ref{fig:hyperparam-sensitivity} show three trends.
First, moderate temporal and representation capacity is enough.
The base window length \(W=30\) and hidden dimension 128 give the best average AUROC and balanced accuracy in their grids.
Larger values do not consistently improve performance.
This suggests that the gain comes from learning the right normal-world structure, not from simply adding capacity.

Second, the number of hyperedges matters.
Eight hyperedges gives the strongest average result across FD002 and FD004.
It reaches average AUROC 0.9807 and average balanced accuracy 0.9345.
Too many hyperedges, such as \(M=32\), hurt FD002 substantially.
This fits the role of the hypergraph branch.
Hyperedges should represent meaningful high-order sensor groups.
An overly large hyperedge set can fragment those relations and make the energy less stable.

Third, context and threshold hyperparameters affect different parts of the score.
For \(\gamma\), the best average AUROC is obtained at 0, while the best average balanced accuracy is obtained at 0.25.
Thus, context modulation is not only a ranking trick.
Its main value is to place the normal boundary correctly under different operating regimes.
The context counterfactual experiment supports this view.
For the threshold quantile, AUROC is unchanged because the ranking energy is fixed.
Balanced accuracy is highest at 0.90.
This confirms that the threshold quantile controls the operating point rather than the learned normal-world representation.

\section{Discussion}

This work is motivated by a common mismatch in real abnormality detection.
Normal behavior is abundant, but abnormal labels are rare and incomplete.
A detector trained to recognize known faults can miss new failure modes.
It can also give a brittle score when the operating context changes.
The proposed normal-world formulation addresses this mismatch by changing the learning target.
The model first learns what normal operation means.
Only after that do a few abnormal examples calibrate the boundary.

This design imposes a clear requirement on the data.
The system should contain repeated normal operation, meaningful temporal structure, and coupled variables.
These assumptions are natural for turbofan engines, industrial processes, clinical monitoring, robotics, traffic systems, and many cyber-physical environments.
In such domains, abnormal events are not arbitrary outliers.
They are often violations of learned dynamics, learned coupling, or learned context compatibility.
This is why a world-model view is useful.
It asks whether an event is coherent under the normal environment, not whether it resembles a small set of labeled failures.

The mechanistic validation experiments are central to this argument.
Unseen-normal acceptance shows that the model does not simply memorize training engines.
Degradation ordering shows that energy has a direction that matches physical deterioration.
Coupling-break and context-counterfactual tests are stronger.
They create samples whose individual parts are still plausible, but whose relations are wrong.
The sharp energy increase under these counterfactual violations suggests that the model has learned part of the normal operating structure.
This evidence is not a complete proof of world understanding, but it is stronger than reporting AUROC alone.
It shows that the learned score responds to structured violations of the normal world.

The results also clarify the role of few-shot abnormal data.
In our formulation, abnormal samples are not used to teach the model all possible faults.
That would be unrealistic when only one or two examples are available.
Instead, they adjust the threshold around an already learned normal region.
This separation is important.
It allows the representation to come from abundant normal data, while the few abnormal examples only decide how conservative the boundary should be.

\section{Limitations}

This paper is an early validation of the normal-world-modeling idea.
The current experiments mainly use C-MAPSS because it is fast, controlled, and suitable for testing whether the proposed energy behaves as expected.
The results support the core hypothesis, but they should not be read as a complete benchmark study across all anomaly-detection domains.
Future work should include more industrial datasets, clinical time-series datasets, and larger real monitoring systems.
Those studies should also use more random seeds and official implementations of recent time-series anomaly detectors under one protocol.

The current implementation is also specialized to multivariate sensor windows.
The broader idea is not limited to sensors.
For video, a normal world may involve consistent object motion, scene dynamics, and physical interactions.
For text, it may involve discourse coherence, factual consistency, and normal semantic transitions.
For images, it may involve compatibility among objects, geometry, and visual context.
Extending the energy formulation from sensor hypergraphs to these modalities is a major direction for future work.

Finally, the present few-shot protocol only calibrates a scalar boundary.
This is a clean starting point, but richer calibration strategies may be useful in practice.
For example, the boundary could be context-dependent, risk-sensitive, or adapted to different costs for false alarms and missed failures.
These extensions would make the method more useful in clinical and industrial deployments.

\section{Conclusion}

We presented a Hypergraph Entropic Normal-World Model for few-shot boundary-calibrated abnormality detection.
The main idea is to reverse the usual abnormal-class learning problem.
Instead of trying to learn a large and diverse failure space from a few labels, the model learns the normal world from abundant normal events.
Abnormality is then scored as departure from this learned world.
A small number of abnormal examples only calibrates the final boundary.

The method combines three sources of normal-world evidence.
The dynamic branch measures temporal surprise.
The hypergraph branch measures high-order sensor consistency.
The manifold branch measures distance from the latent normal region.
Together, these terms define an entropy-aware energy that can be used as an anomaly score, a normality score, and a risk margin.
On C-MAPSS, the full energy performs strongly in both zero-shot and calibrated settings.
The gains are most clear on FD004, where multiple operating conditions and fault modes make simple marginal scoring unreliable.

The mechanistic tests add a second layer of evidence.
The learned energy accepts unseen healthy engines, increases along degradation trajectories, and reacts strongly to coupling and context violations.
These tests support the central claim that the model learns structured normal-world knowledge rather than only an input-output mapping.
Although the present study is still a verification-stage experiment, it points to a general strategy for abnormality detection under label scarcity.
Learning the normal world first may provide a practical foundation for fault detection, clinical monitoring, and other domains where abnormal labels are rare but normal experience is abundant.

\end{document}